\documentclass{article}

\PassOptionsToPackage{numbers, compress}{natbib}

\usepackage[final]{neurips_2020}

\usepackage[utf8]{inputenc} 
\usepackage[T1]{fontenc}    
\usepackage{hyperref}       
\usepackage{url}            
\usepackage{booktabs}       
\usepackage{amsfonts}       
\usepackage{nicefrac}       
\usepackage{microtype}      
\usepackage{graphicx}
\usepackage{kotex}
\usepackage{subcaption}
\usepackage{array}
\usepackage{multirow}

\newcolumntype{C}[1]{>{\centering\let\newline\\\arraybackslash\hspace{0pt}}m{#1}}
\def\FootNoteVersion{0}

\title{Human-Like Active Learning:\\ Machines Simulating the Human Learning Process}

\author{Jaeseo Lim\thanks{Equal contribution $\dagger$ Shared corresponding authors}$^{\ 1}$ \qquad Hwiyeol Jo\footnotemark[1]$\ ^{2,3}$ \qquad Byoung-Tak Zhang\footnotemark[2]$\ ^{1,3,4}$ \qquad Jooyong Park\footnotemark[2]$\ ^{1,5}$ \\
  $^{1}$Interdisciplinary Program in Cognitive Science, Seoul National University \\
  $^{2}$Institute of Computer Technology, Seoul National University \\
  $^{3}$Department of Computer Science and Engineering, Seoul National University \\
  $^{4}$SNU AI Institute, Seoul National University \\
  $^{5}$Department of Psychology, Seoul National University \\
  {\tt \{jaeseolim,jooypark\}@snu.ac.kr} \quad {\tt hwiyeolj@gmail.com} \quad {\tt btzhang@bi.snu.ac.kr} \\
}

\begin{document}

\maketitle

\begin{abstract}
    Although the use of active learning to increase learners' engagement has recently been introduced in a variety of methods, empirical experiments are lacking. In this study, we attempted to align two experiments in order to (1) make a hypothesis for machine and (2) empirically confirm the effect of active learning on learning. 
    In Experiment 1,we compared the effect of a passive form of learning to active form of learning. The results showed that active learning had a greater learning outcomes than passive learning. In the machine experiment based on the human result, we imitated the human active learning as a form of knowledge distillation. The active learning framework performed better than the passive learning framework. In the end, we showed not only that we can make build better machine training framework through the human experiment result, but also empirically confirm the result of human experiment through imitated machine experiments; human-like active learning have crucial effect on learning performance.
\end{abstract}

\section{Introduction}
The current educational environment often utilizes passive teaching methods that simply delivers information since it requires students to learn a large amount of knowledge at a limited amount of time. Although passive learning have the advantage of being able to deliver a lot of knowledge, such characteristic this does not directly lead to learners' achievement. Rather, there are many studies that show the problems of passive form of learning.

Psychologists have perceived that, although learners can learn from receiving knowledge passively, they perform much better by learning actively. Active learning
\if\FootNoteVersion\footnote{Note that the term `active learning' is differently used in psychology and computer science.}\fi
is defined by educational researchers as learning that requires students to engage cognitively and meaningfully with the learning materials~\cite{bonwell1991active}. As students become more active in learning, they get to move in class, or really think about what they learn by analyzing, synthesizing, and evaluating materials rather than just passively receiving it~\cite{chi2014icap, corno1983role}.

In this paper, the advantages of active learning are outlined, along with the problems of passive learning. Furthermore, through human-like active learning in machine experimentation, we empirically explored the benefits of active learning. 

\paragraph{The Necessity of Active Learning}
As an alternative to passive learning, various methods have been researched to increase learners' participation. They are called active learning, which requires learner’s cognitive intervention~\cite{bonwell1991active}. According to \cite{menekse2013differentiated}, the main constructs of active learning are students’ engagement with concrete learning experiences, knowledge construction through meaningful activities, and some degree of interaction between students during the learning process. Therefore, active learning eagers innovative learner-centered instructional approach that dynamically involves learners in the learning process. 

As a segmentation for active learning, Chi and colleagues \cite{chi2014icap, chi2009active} proposed the `Interactive-Constructive-Active-Passive (ICAP)' framework. The ICAP framework classifies active learning into three stages, interactive, constructive, and active, according to the learner’s level of cognitive engagement. The passive mode generally refers to a situations where learners listen to lectures, while in active modes learners physically manipulate the information such as learning materials in educational settings. In constructive modes, learners make better efforts to gain knowledge and proceed with the action of making the study material their own by drawing diagrams or asking questions. In interactive modes, two or more colleagues cooperate and co-construct through the process of asking questions and responding to one another during their conversation. Therefore, learners' academic achievement was lowest at P, then increased at A, C, and I in the ascending order. These research demonstrate that active learning, when used appropriately, can enhance learning to a greater extent than passive learning performed in the same amount of time.~\cite{lim2019active}\\
\paragraph{Present Study: Human-like Active Learning}
In recent years, active learning has also been frequently used in machines
~\r Active learning in machine learning which query the datasets to be labeled for training by an oracle may get higher accuracy. Usually, most active learning in machine learning method focused on mechanism for choosing queries, or only on its high performance. In other words, it has become a learning method for machines, not essentially human's active learning.

This study aims to identify the effectiveness of active learning based on the ICAP framework and its impact on learners' learning performance. Accordingly, we compared performances of students who learn actively with those of students who learn passively. Thus, the lecture group was set up as a condition for passive learning. On the other hand, the discussion groups was set up as a condition for active learning.

On the next step, we simulate results of human-like active learning  by using machine learning. The machine can complement the limitation of human experiment such as sampling bias, and human subjectivity. Therefore, we intended to maximize the effectiveness of human experimentation through the validation of machine experiments. Therefore, in order to form a form of active learning in human, we have set up teacher models and student models.



\section{Experiment 1: Humans} 
Experiment 1 sought to find out which learning method produces better performance. Here, passive learning was defined as listening to lectures, a traditional learning method, whereas active learning was set as engaging in discussions. 

\subsection{Methodology}
{\bf Participants and Design.}  Fifty-four undergraduate students in selective university participated in this experiment. Participants were assigned to each group randomly: the lecture group ({\tt L} group, \textit{n}=25), a passive form of learning and the discussion group ({\tt D} group, \textit{n}=29, $\#\texttt{groups}$=9), a active form of learning. Three or four students formed a discussion group.\\
{\bf Procedure.} The participants first took a background knowledge questionnaire. The {\tt L} groups watched the video lecture and studied the provided written learning material by themselves without any physical manipulation for 36 minutes. Students of {\tt D} groups studied the written learning materials by themselves for 18 minutes and then, discussed in groups of three or four for another 18 minutes. In fact, the total amount of learning time for both groups was the same. Lastly, all two groups took a 20 minutes test.

\subsection{Result and Discussion}
\begin{figure*}[t]\centering
    \includegraphics[scale=0.3]{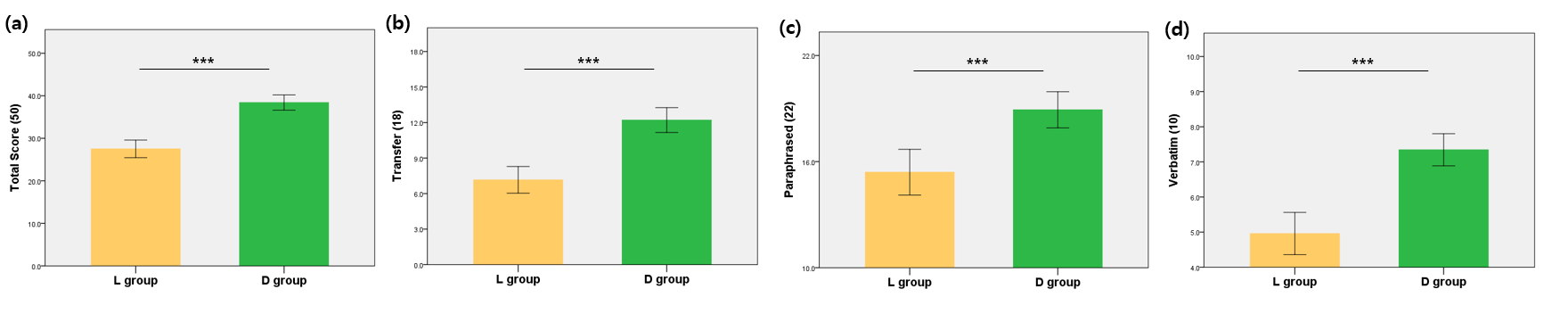}
    \caption{
Mean scores for the total and the three different question types. (a) total score;  (b) transfer type questions; (c) paraphrased type questions; (d) verbatim type questions. Gender and age were adjusted. ***$ps<.001$. Error bars indicate ±2$SE$.}\label{fig:study1}
\end{figure*}

Analysis of covariance (ANCOVA) was conducted to examine the differences between the two groups for the three type of test questions (see Appendix). The results revealed that the total means of the {\tt D} group ({\small $M=38.41$, $SD=4.74$}) was significantly higher than that of {\tt L} group ({\small $M=27.52$, $SD=5.03$}), {\small $F(1,52)=61.31$, $p< .001$, ${\eta}^2= .55$}. For the transfer type items, the {\tt D} group ({\small $M=12.21$, $SD=2.76$}) scored higher than the {\tt L} group ({\small $M=7.16$, $SD=2.73$}), {\small $F(1,52)=42.62$, $p<.001$, ${\eta}^2 =.46$}. For the paraphrased type questions, the {\tt D} group ({\small $M= 18.93$, $SD=3.14$}) scored significantly higher than the {\tt L} group ({\small $M=15.40$, $SD=2.69$}), {\small $F(1,52) = 15.74$, $p<.001$, ${\eta}^2 = .24$}. Lastly, for the verbatim type questions, the {\tt D} group ({\small $M= 7.35$, $SD= 1.20$}) scored significantly higher than the {\tt L} group ({\small $M=4.96$, $SD=1.46$}), {\small $F(1,52)=40.60$, $p<.001$, ${\eta}^2 = .45$}. The average and standard deviation of the test scores are provided in Figure~\ref{fig:study1}. 

In line with our hypothesis, the {\tt D} group scored much higher than the {\tt L} group in all of test question types. Discussions, active learning, promoted greater learning outcome than lectures, passive learning.  Consistent with the ICAP framework, the findings showed the learning benefits of active learning. Subsequently, we compared active and passive learning in machines, in order to further validate the out results of human-like active learning. 



\section{Experiment 2: Machines}
\subsection{Methodology}
{\bf Datasets and Classifiers. }
    We used five publicly open text classification datasets.\footnote{Experiment 1 is actually open-ended QA tasks, but for simplicity we use the basic tasks.} Three are topic classification datasets: DBpedia ontology (DBpedia)~\cite{lehmann2015dbpedia}, YahooAnswers (Yahoo)~\cite{chang2008importance}, AGNews and the other two are sentiment classification datasets: Yelp reviews (Yelp)~\cite{zhang2015character}, IMDB~\cite{maas2011learning}.\\
    Next, we used TextCNN~\cite{kim2014convolutional} and LSTM~\cite{article} as classifiers, but we made a difference on model capacity between passive learning and active learning. Passive learning required a teacher model ($M_{T}$), which is able to learn from the data fully. On the other hand, student models ($M_{S}$) only represent novice learners. Thus, $M_T$ has more deep and complex architecture whereas $M_S$ has shallow and simple architecture. The details of architecture of TextCNN and LSTM will be described in Appendix. We optimize their loss (see Figure~\ref{fig:2}) using Adam~\cite{kingma2014adam}. The other hyperparameters (e.g., learning rate=1e-3, batch size=64) were the same.\\
{\bf Implementation. }
    Knowledge transfer was implemented by knowledge distillation~\cite{hinton2015distilling}. The method did not use training data directly but it used other models to train a model. That is, a model can learn the other models' prediction scores on the training data. The transfer is implemented by mean squared error loss between two model predictions. By using the idea, the training frameworks were illustrated in Figure~\ref{fig:2}. Passive learning (b) used a teacher model ($M_T$) and a student model ($M_S$). Both models are trained on the conventional training framework (see (a)) and then knowledge transfer occurred from $M_T$ to $M_S$; it imitates ``the teacher provides knowledge to the student.'' Lastly, (c) imitates active learning used in Experiment 1. Limiting the time (36 mins in passive learning vs. 18 mins in active learning) in Experiment 1 corresponds to constraint the training capacity for machine. Therefore, we used two $M_S$, which is much smaller than $M_T$. Beside, in order to implement discussion, we simply made the knowledge transfer (distillation) in bidirectional ways. Overall method in (c) then imitates that ``students use their knowledge (inter)actively to make better results.''

\begin{figure*}[t] \centering
    \begin{subfigure}{0.19\textwidth}\centering
    \caption{}
    \includegraphics[scale=0.31]{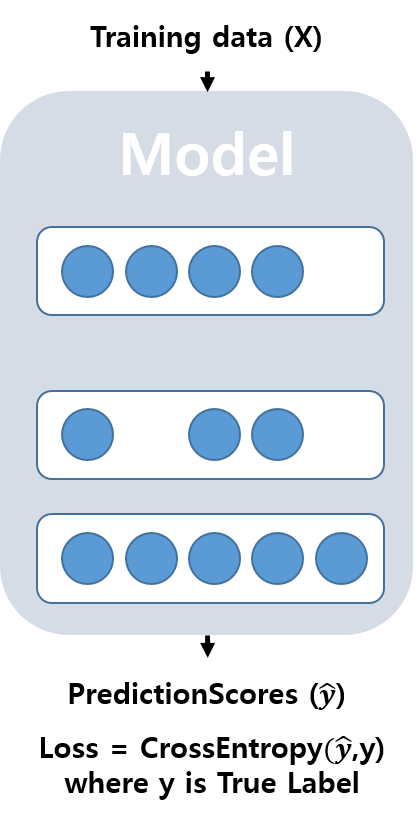}
    \end{subfigure}
    \begin{subfigure}{0.36\textwidth}\centering
    \caption{}
    \includegraphics[scale=0.31]{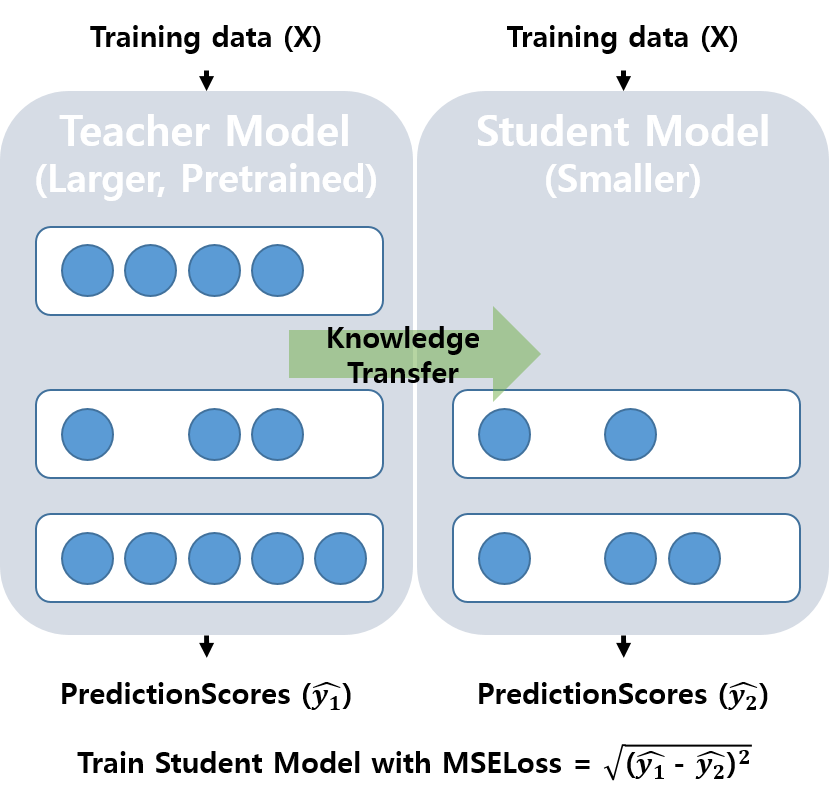}
    \end{subfigure}
    \begin{subfigure}{0.36\textwidth}\centering
    \caption{}
    \includegraphics[scale=0.31]{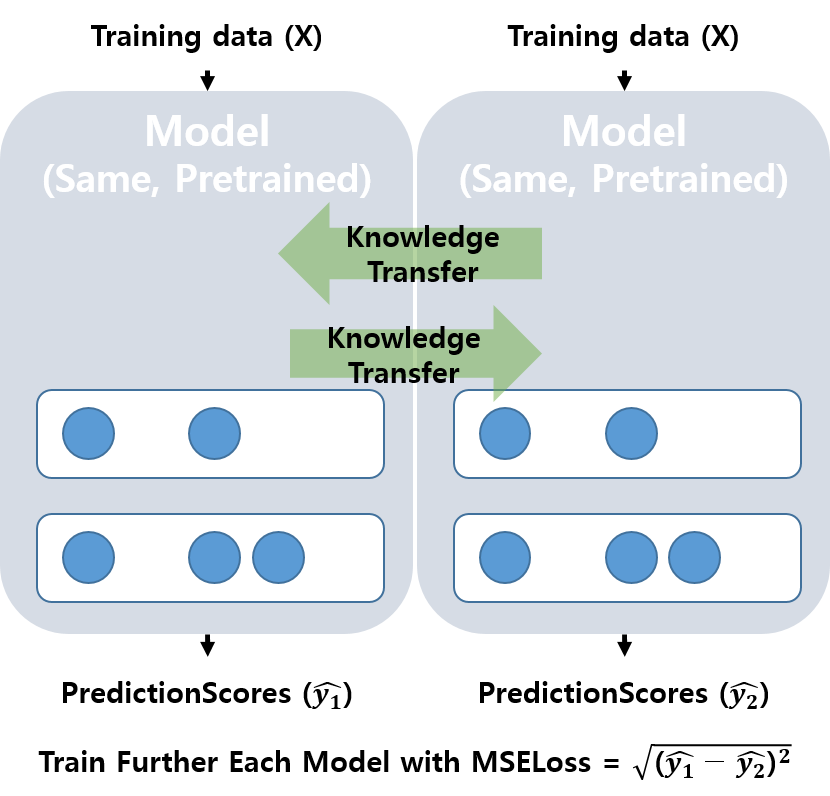}
    \end{subfigure}
    \caption{Illustration of the training frameworks. (a) the conventional training framework.
    {\small\tt Pretrained} means the model is trained like (a) before using it. (b) passive learning using knowledge distillation (in this case, transfer), which minimizes the loss between the prediction scores of the two models instead of the loss between the prediction scores and the true labels.
    (c) implementation to imitate active learning.}\label{fig:2} 
\end{figure*}

\begin{table*}[h!]
    \small
    \begin{center} 
    \caption{The performance of training frameworks on the text classification. $x$ denotes training data, $M_T$ denotes the teacher model, which had larger model capacity than the student model ($M_S$).
    The arrows ($\rightarrow$ and $\leftrightarrow$) describe the flow of knowledge transfer.
    Passive and Active can be symbolized as $[x \rightarrow M_T] \rightarrow M_S$ and $[x \rightarrow M_S] \leftrightarrow [x \rightarrow M_S]$, respectively.}
    \label{tab:1}
    \vskip 0.12in
    \begin{tabular}{C{1.5cm}|c|ccccc} 
    \hline
    \bf Classifier & \bf Methods & IMDB & Yelp & AGNews & Yahoo & DBpedia \\
    \hline
    \multirow{4}{*}{CNN} & $x \rightarrow M_T$ & 76.61$\pm$.73 & 56.36$\pm$.17 & 88.85$\pm$.38 & 65.45$\pm$.30 & 98.04$\pm$.11 \\
    \cline{2-7}
    & $x \rightarrow M_S$ & 78.70$\pm$.13 & 56.31$\pm$.25 & 89.54$\pm$.12 & 67.92$\pm$.25 & 98.01$\pm$.03 \\
    \cline{2-7}
    & Passive & 78.89$\pm$.37 & 56.60$\pm$.08 & 89.68$\pm$.28 & 66.01$\pm$.36 & 97.85$\pm$.12 \\
    \cline{2-7}
    & Active & \bf 79.04$\pm$.28 & \bf 56.79$\pm$.15 & \bf 90.21$\pm$.13 & \bf 68.69$\pm$.10 & \bf 98.14$\pm$.03\\
    \hline
    \multirow{4}{*}{LSTM} & $x \rightarrow M_T$ & 77.05$\pm$.13 & 58.94$\pm$.19 & 89.38$\pm$.34 & 72.23$\pm$.20. & 98.43$\pm$.05 \\
    \cline{2-7}
    & $x \rightarrow M_S$ & 77.10$\pm$.25 & 58.26$\pm$.24 & 89.45$\pm$.47 & 71.63$\pm$.75 & 98.26$\pm$.06 \\
    \cline{2-7}
    & Passive & 77.55$\pm$.82 & 58.90$\pm$.20 & 89.74$\pm$.06 & 72.93$\pm$.78 & 98.33$\pm$.02 \\
    \cline{2-7}
    & Active & \bf 77.58$\pm$.16 & \bf 59.00$\pm$.14 & \bf 90.53$\pm$.23 & \bf 74.44$\pm$.55 & \bf 98.67$\pm$.06 \\
    \hline
    \end{tabular}
    \end{center} 
\end{table*}
{\bf Result and Discussion}
    The performance of the passive learning framework and the active learning framework are presented in Table~\ref{tab:1}. When we compared the result between the passive and the active, the active learning performed better in most of the datasets. These results support our hypothesis that active learning enhances performance, as observed in Experiment 1. We also investigated that the performance of the passive learning framework were on par with the conventional learning framework ($x \rightarrow M_S$), and even were better on several datasets. The reason might be that the teacher model enabled to capture a higher level of representation, and the knowledge is transferred to the student model. Moreover, in some datasets the teacher model might be overfitted to the training data, so their performance on the test data was worse than the student model.

\section{Conclusions}


In this study, we conducted two experiments in order to investigate the effect of active learning on performance. Active learning are generally expected to enhance learners' performance better than passive learning. Because actively participating in learning process allows the learners to activate relevant knowledge, thereby allowing the learners to assimilate novel information to fill in the knowledge gaps, whereas passive learning only allows to store novel information for a while~\cite{menekse2013differentiated}. With this expectation, in Experiment 1, we compared two conditions: lecture (passive learning) and discussion (active learning). As a result, the discussion group scored higher than the lecture group in all types of questions, as expected. These findings also correspond with the ICAP framework that learning performance would be greater in active learning than in passive learning. 

In Experiment 2, we compared performance of active learning with passive learning in machines. Like in the human experiment, machines also increased their performance when they performed human-like active learning. In other words, the two student models exchanged opinions was more efficient than the well-learned teacher model transferring knowledge. We believe that these cognitive processes based approach would help the researchers to build better architectures.

\bibliographystyle{plainnat}
\bibliography{neurips_2020}

\newpage
\appendix

\section{Appendix. Samples of three types of test questions: verbatim, paraphrased, and transfer items}

(1) Examples of verbatim item: Given that there is no one who filed an accusation against a crime subject for prosecution, prosecutors must designate a person who can file the complaint within (   ) days upon the request of the stakeholders. \\

(Answer: 10) \\

(2) Examples of paraphrased items: Explain who the entitled person with the right to file a complaint is. \\

(Answer: Provided that there is no one to make the accusation (in case of an offense subject to complaint), prosecutors shall designate the person with the right to file a complaint within 10 days upon request by stakeholders) \\

(3) Examples of transfer items: 17. The under-aged victim (V) accused the offender (D) of contempt, and then withdrew his accusation on July 26th, 2017. Afterwards V’s mother (M), the legal representative of V, accused D on August 3rd, 2017. D was charged with contempt and was found guilty on the first trial. However, D made an appeal claiming M’s complaint is not valid because V has already withdrawn his complaint, and thus, the prosecutor’s indictment is against the provisions of the law. Will the Court of Appeals accept D’s claim? \\

(Answer: A legal representative of an under-aged victim can independently file a complaint regardless of whether the victim’s complaint is nullified. Such complaint can even go against the victim’s stated will. Thus, even if victim V withdraws his accusation, the complaint of V’s legal representative M is still effective. In conclusion, the Court of Appeals will reject D’s claim)

\section{Appendix. The details of Teacher Model and Student Model}
    In TextCNN, teacher model ($M_T$) consisted of 2 convolution layers, which had 32 and 16 channels, respectively. We also utilized multi-kernel approaches, which kernel sizes were 2, 3, 4, and 5. On the other hand, student model $M_S$ consisted of 1 convolution layer, which had 32 channels only. Moreover, its kernel size were 2 and 3 only.\\
    Likewise, in LSTM, the teacher model architecture consisted of forward and backward LSTM layers (i.e., bidirectional) with 300 hidden nodes. In contrast, the student model architectures had forward LSTM layers only with 150 hidden nodes.


\end{document}